\documentclass{llncs}
\usepackage{llncsdoc}
\usepackage{cite}
\usepackage{graphicx}
\usepackage{upgreek}
\usepackage[cmex10]{amsmath}
\usepackage{amsfonts}
\usepackage{amssymb}
\usepackage{enumerate}
\usepackage{url}

\usepackage{multirow}
\usepackage{scrextend}
\usepackage{tabularx}
\usepackage{times}
\usepackage{fixltx2e}
\usepackage{subfigure}
\usepackage{floatrow}

\newcommand*\samethanks[1][\value{footnote}]{\footnotemark[#1]}

\begin{document}

\title{DeepLesion: Automated Deep Mining, Categorization and Detection of Significant Radiology Image Findings using Large-Scale Clinical Lesion Annotations}


\author{Ke Yan\samethanks[1] \and Xiaosong Wang\thanks{These two authors contributed equally.} \and Le Lu \and Ronald M. Summers}

\institute{Department of Radiology and Imaging Sciences,\\ National Institutes of Health Clinical Center, Bethesda, MD 20892 \\ \email{\{ke.yan, xiaosong.wang, le.lu, rms\}@nih.gov} }

\maketitle
\begin{abstract}
Extracting, harvesting and building large-scale annotated radiological image datasets is a greatly important yet challenging problem. It is also the bottleneck to designing more effective data-hungry computing paradigms (e.g., deep learning) for medical image analysis. Yet, vast amounts of clinical annotations (usually associated with disease image findings and marked using arrows, lines, lesion diameters, segmentation, etc.) have been collected over several decades and stored in hospitals' Picture Archiving and Communication Systems. In this paper, we mine and harvest one major type of clinical annotation data -- lesion diameters annotated on bookmarked images -- to learn an effective multi-class lesion detector via unsupervised and supervised deep Convolutional Neural Networks (CNN). Our dataset is composed of 33,688 bookmarked radiology images from 10,825 studies of 4,477 unique patients. For every bookmarked image, a bounding box is created to cover the target lesion based on its measured diameters. We categorize the collection of lesions using an unsupervised deep mining scheme to generate clustered pseudo lesion labels. Next, we adopt a regional-CNN method to detect lesions of multiple categories, regardless of missing annotations (normally only one lesion is annotated, despite the presence of multiple co-existing findings). Our integrated mining, categorization and detection framework is validated with promising empirical results, as a scalable, universal or multi-purpose CAD paradigm built upon abundant retrospective medical data. Furthermore, we demonstrate that detection accuracy can be significantly improved by incorporating pseudo lesion labels (e.g., Liver lesion/tumor, Lung nodule/tumor, Abdomen lesions, Chest lymph node and others). This dataset will be made publicly available (under the open science initiative).
\end{abstract}

\section{Introduction}
\label{intro}
Computer-aided detection/diagnosis (CADe/CADx) has been a highly prosperous and successful research field in medical image processing. Many commercial software packages have been developed for clinical usage and screening. Recent advances (e.g., automated classification of skin lesions~\cite{esteva2017dermatologist}, detection of liver lesion \cite{Green16Liv}, pulmonary embolism \cite{Taj15Pulmonary}) have attracted even more attention to the application of deep learning paradigms to CADe/CADx. Deep learning, namely Convolutional Neural Network (CNN) based algorithms, perform significantly better than conventional statistical learning approaches combined with hand-crafted image features. However, these performance gains are often achieved at the cost of requiring tremendous amounts of training data accompanied with high quality labels. Unlike general computer vision tasks, medical image analysis currently lacks a substantial, large-scale annotated image dataset (comparable to ImageNet \cite{deng2009imagenet} and MS COCO \cite{lin2014microsoft}),for two main reasons: 1) The conventional methods for collecting image labels via Google search + crowd-sourcing from average users cannot be applied in the medical image domain, as medical image annotation reuqires extensive clinical expertise; 2) Significant inter and intra-observer variability (among even well-trained, experienced radiologists) frequently occurs, and thus may compromise reliable annotation of a large amount of medical images, especially considering the great diversity of radiology diagnosis tasks.

Current CADe/CADx methods generally target one particular type of diseases or lesions, such as lung nodules, colon polyps or lymph nodes \cite{Liu16Rcnn}. Yet, this approach differs from the methods radiologists routinely apply to read medical image studies and compile radiological reports. Multiple findings can be observed and are often correlated. For instance, liver metastases can spread to regional lymph nodes or other body parts. By obtaining and maintaining a holistic picture of relevant clinical findings, a radiologist will be able to make a more accurate diagnosis. However, it remains greatly challenging to develop a universal or multi-purpose CAD framework, capable of detecting multiple disease types in a seamless fashion. Such a framework is crucial to building an automatic radiological diagnosis and reasoning system.  



\begin{figure}
	\centering
	\subfigure[]{\includegraphics[width = 0.32\textwidth, trim=40 40 0 20mm, clip]{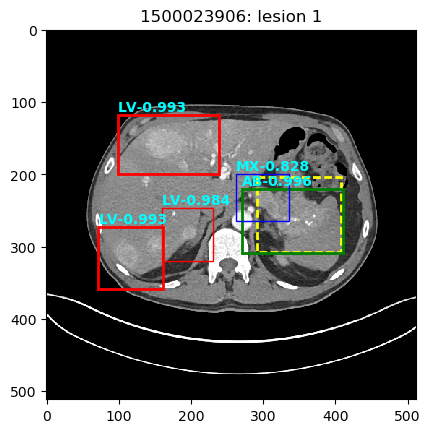}}
	\subfigure[]{\includegraphics[width = 0.32\textwidth, trim=40 27 0 25mm, clip]{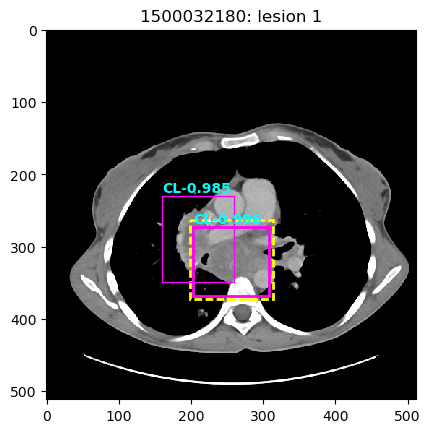}}
	\subfigure[]{\includegraphics[width = 0.32\textwidth, trim=40 55 0 15mm, clip]{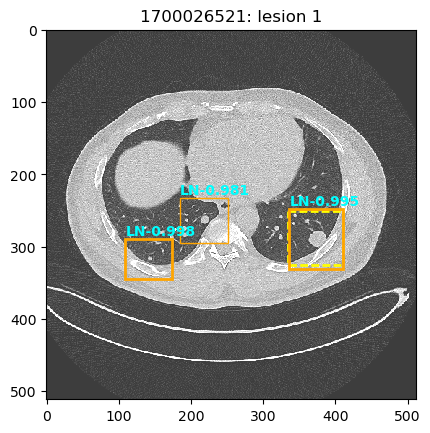}} 
	\caption{Three sample bookmarked images illustrated with annotation lesion patches (i.e., yellow dashed boxes). The outputs from our proposed multi-category lesion detection framework are shown in colored boxes with LiVer lesion (LV) in Red, Lung Nodule (LN) in Orange, ABdomen lesion (AB) in Green, Chest Lymph node (CL) in magenta and other MiXed lesions (MX) in blue. (a) A spleen metastasis is correctly detected along with several liver and abdomen metastases; (b) Two large lymph nodes in mediastinum are all correctly detected; (c) All three lung nodules are detected despite two small ones not being annotated in this bookmarked image.  
	}
	\label{fig:hightlight_results} \vspace{-3mm}
\end{figure}

In this paper, we attempt to address these challenges by first introducing a new large-scale dataset of bookmarked radiology images, which accommodate lesions from multiple categories. Our dataset, named DeepLesion, is composed of 33,688 bookmarked images from 10,825 studies of 4,477 patients (see samples in Fig. \ref{fig:hightlight_results}). For each bookmarked image, a bounding box is generated to indicate the location of the lesions. Furthermore, we integrate an unsupervised deep mining method to compute {\em pseudo} image labels for database self-annotating. Categories of Liver lesion/tumor, Lung nodule/tumor, Abdomen lesions, Chest lymph node and others are identified by our computerized algorithm instead of radiologists' annotation, which may be infeasible. After obtaining the dataset, we develop an automatic lesion detection approach to jointly localize and classify lesion candidates using discovered multiple categories. Last, how the unsupervisedly-learned pseudo lesion labels affect the deep CNN training strategies and the quantitative performance of our proposed multi-class lesion detector is investigated.
 
\section{Methods}
\label{method}
In this section, we first describe how our DeepLesion dataset is constructed. Next, we propose an unsupervised deep learning method to mine the latent lesion categories in each image. This method involves an iterative process of deep image feature extraction, image clustering and CNN model retraining. Finally, we present a multi-class object detection approach to detect lesions of multiple categories. 


\subsection{DeepLesion Dataset} \label{database}
Radiologists routinely annotate hundreds of clinically meaningful findings in medical images, using arrows, lines, diameters or segmentations to highlight and measure different disease patterns to be reported. These images, called ``bookmarked images'', have been collected over close to two decades in our institute's Picture Archiving and Communication Systems (PACS). Without loss of generality, in this work, we study one type of bookmark in CT images: lesion diameters. Each pair of lesion diameters consists of two lines, one measuring the longest diameter and the second measuring its longest perpendicular diameter in the plane of measurement. We extract the lesion diameter coordinates from the PACS server and convert into  corresponding positions in the image plane coordinates, noted as $\{(x_{11},y_{11}),(x_{12},y_{12})\};\{(x_{21},y_{21}),(x_{22},y_{22})\}$. A bounding box $(left_x, top_y, width,$ $ height)$ is computed to cover a rectangular area enclosing the lesion measurement with 20 pixel padding in each direction, i.e., $(x_{min}-20, y_{min}-20, x_{max}-x_{min}+40, y_{max}-y_{min}+40)$ where $x_{min} = Min(x_{11},x_{12},x_{21},x_{22})$ and $x_{max} = Max(x_{11},x_{12},x_{21},x_{22})$, and similarly for $y_{min}$ and $y_{max}$. The padding range can capture the lesion's full spatial extent with sufficient image context. We thus generate 33,688 bookmarked radiology images from 10,825 studies of 4,477 unique patients, and each bookmarked image is associated with a bounding box annotation of the enclosed lesion. Sample bookmarked images and bounding boxes are shown in Fig.~\ref{fig:hightlight_results}. 
  
\subsection{Unsupervised Lesion Categorization}
\label{annot}
The images in our constructed lesion dataset contain several types of lesions commonly observed by radiologists, such as lung nodule/lesion, lymph node, and liver/kidney lesion. However, no detailed precise category labels for each measured lesion have been provided. Obtaining such from radiologists would be highly tedious and time-consuming, due to the vast size and comprehensiveness of DeepLesion. To address this problem, we propose a looped deep optimization procedure for automated category discovery, which generates visually coherent and clinically-semantic image clusters. Our algorithm is conceptually simple: it is based on the hypothesis that the optimization procedure will ``converge'' to more accurate labels, which will lead to better trained CNN models. Such models, in turn, will generate more representative deep image features, which will allow for creating more meaningful lesion labels via clustering. 

\begin{figure}[t]
	\centering
	\includegraphics[width=1.00\linewidth]{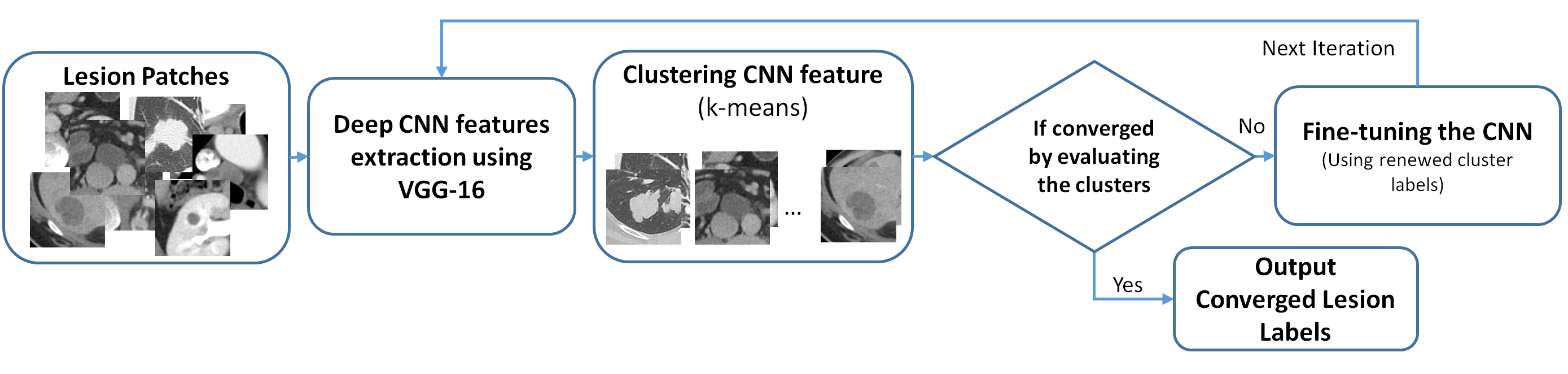}
	\caption{Lesion categorization framework via unsupervised and iteratively-optimized deep CNNs.}
	\label{fig:LDPO_framework:png} \vspace{-3mm}
\end{figure}

As a pre-processing step, we crop the lesion patches from the original DICOM slides using the dilated bounding boxes (described in Sec. \ref{database}) and resize them, prior to feeding them into the CNN model. As shown in Fig.~\ref{fig:LDPO_framework:png}, our iterative deep learning process begins by extracting deep CNN features for each lesion patch using the ImageNet \cite{deng2009imagenet}) pre-trained VGG-16 \cite{Sim15Vgg} network. Next, it applies $k$-means clustering to the deep feature encoded lesion patches after $k$ is determined via model selection \cite{Gomes2010}. Next, it fine-tunes the current VGG-16 using the new image labels obtained from $k$-means. This yields an updated CNN model for the next iteration. The optimization cycle  terminates once the convergence criteria have been satisfied. 

{\bf Encoding Lesion Patches using Deep CNN Features:} The VGG-16 \cite{Sim15Vgg} CNN architecture is adopted for patch encoding and CNN model fine-tuning to facilitate the iterative procedure. The image features extracted from the last fully-connected layer (e.g., FC6/FC7 of VGG-16) are used, as they are able to capture both the visual appearance  and the spatial layout of any lesion, with its surrounding context.

{\bf Convergence in Patch Clustering and Categorization:} We hypothesize that the newly generated clusters will converge to the ``oracle'' label clusters, after undergoing several staged of cluster optimization. Two convergence measurements are employed: Purity~\cite{Gomes2010} and Normalized Mutual Information (NMI). We assess both criteria by computing empirical similarity scores between clustering results from two adjacent iterations. If the similarity score exceeds a pre-defined threshold, the optimal clustering driven categorization of lesion patches has been attained. For each iteration, we randomly shuffle the lesion patches and divide the data into three subsets: training (75\%), validation (10\%) and testing (15\%). Therefore the ``improving-then-saturating'' trajectory of the CNN classification accuracy on the testing set can also indicate the convergence of the clustering labels (i.e., optimal image labels have been obtained). 

\subsection{Multi-category Lesion Detection}
\label{faster}

\begin{figure}[t]
	\
	\includegraphics[width=0.98\linewidth]{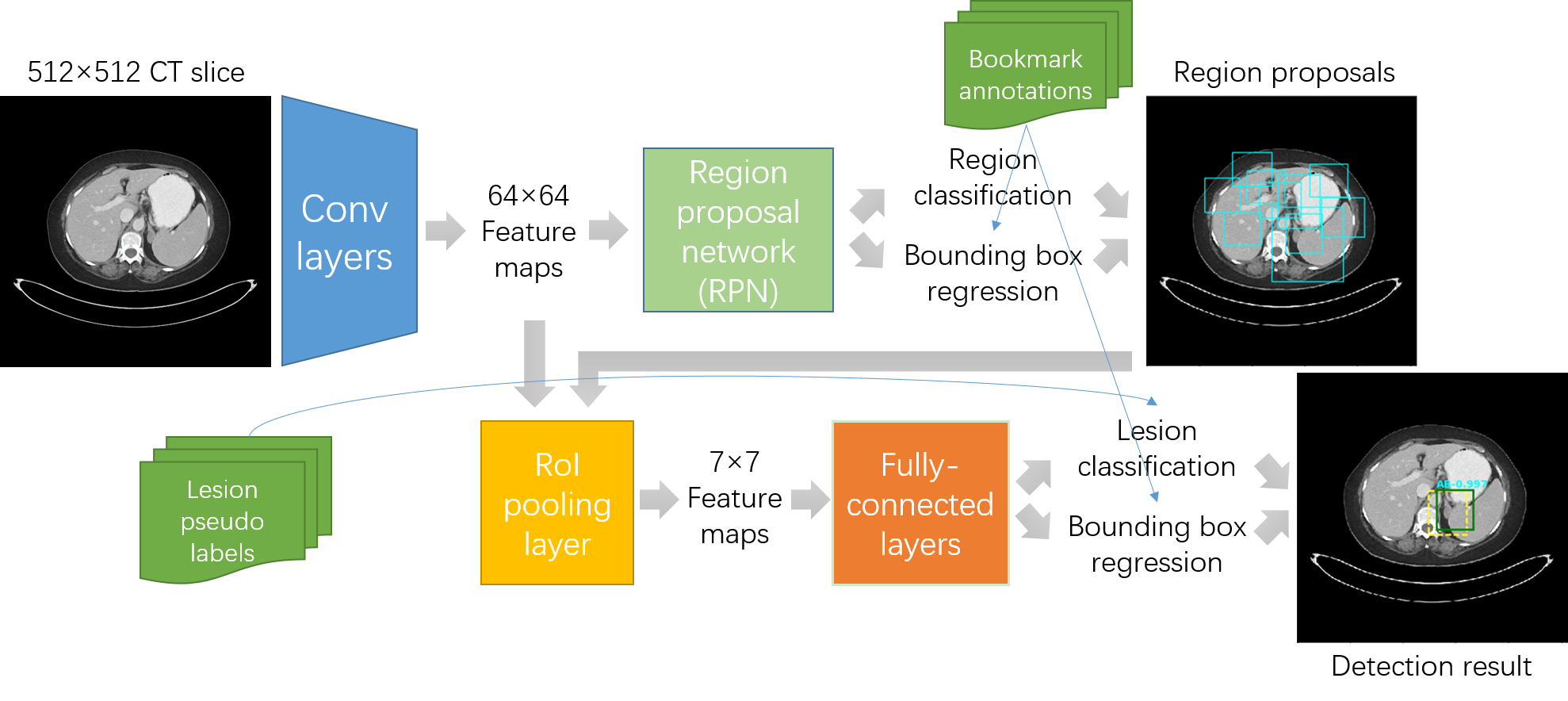} 
	\caption{Flow chart of the lesion detection algorithm. Bookmarked clinical annotations provide the ground-truth bounding boxes of lesions for detector training. In detection, the dashed and solid boxes indicate the ground-truth annotation and its predicted lesion detection, respectively.}
	\label{fig:det_framework} \vspace{-4mm}
\end{figure}

Using the bounding boxes (Sec. \ref{database}) and their corresponding newly generated pseudo-category labels (Sec. \ref{annot}), we develop a multi-class lesion detector adapted from the Faster RCNN method \cite{Ren15Faster}. An input image is first processed by several convolutional and max pooling layers to produce feature maps, as shown in Fig.\ \ref{fig:det_framework}. Next, a region proposal network (RPN) parses the feature maps and proposes candidate lesion regions. It estimates the probability of ``target/non-target'' on a fixed set of anchors (candidate regions) on each position of the feature maps. Furthermore, the location and size of each anchor are fine-tuned via bounding box regression. Afterwards, the region proposals and the feature maps are sent to a Region of Interest (RoI) pooling layer, which re-samples the feature maps inside each proposal to a fixed size (we use 7$ \times $7 here). These feature maps are then fed into several fully-connected layers that predict the confidence scores for each lesion class and run another bounding box regression for further fine-tuning. Non-maximum suppression (NMS) is then applied to the feature maps. Finally, the system returns up to five detection proposals with the highest confidence scores ($>0.5$), as each image only has one bookmarked clinical annotation.

The ImageNet pretrained VGG-16 \cite{Sim15Vgg} model is adopted as the backbone of Faster RCNN \cite{Ren15Faster}. It is useful to remove the last pooling layer (pool4) in VGG-16 to enhance the resolution of the feature map and to increase the sampling ratio of positive samples (candidate regions that contain lesions). In our experiments, removing pool4 improves the accuracy by $\sim$ 15\%. It is critical to set the anchor sizes in RPN to fit the size of ground-truth bounding boxes in DeepLesion dataset. Hence, we use anchors of three scales (48, 72, 96) and aspect ratios (1:1, 1:2, 2:1) to cover most of the boxes.

For image preparation, we use the ground-truth lesion bounding boxes derived in Sec. \ref{database} incorporating enlarged spatial contexts. Each full-slice image in the detection phase is resized, so that the longest dimension is of 512 pixels. We then train the network as demonstrated in Fig. \ref{fig:det_framework} in a multi-task fashion: two classification and two regression losses are jointly optimized. This end-to-end training strategy is more efficient than the four-step method in the original Faster RCNN implementation \cite{Ren15Faster}. During training, each mini-batch has 4 images, and the number of region proposals per image is 32. We use the Caffe toolbox and the Stochastic Gradient Descent (SGD) optimizer. The base learning rate is set to 0.001, and is reduced by a factor of 10 every 20K iterations. The network generally converges within 60K iterations. \vspace{-2mm}

\section{Results and Discussion}
\label{result}

\begin{figure}[t]
	\centering
	\subfigure[]{\includegraphics[width = 0.300\textwidth, trim=35 60 0 17mm, clip]{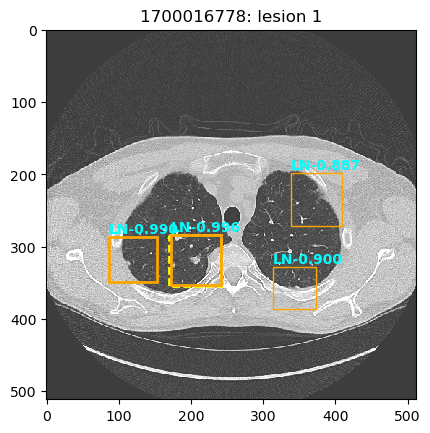}}
	\subfigure[]{\includegraphics[width = 0.300\textwidth, trim=35 65 0 15mm, clip]{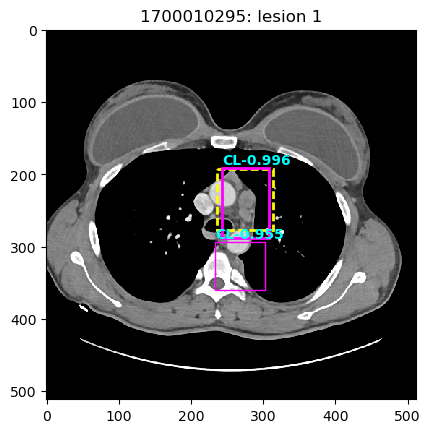}}
	\subfigure[]{\includegraphics[width = 0.300\textwidth, trim=35 50 0 20mm, clip]{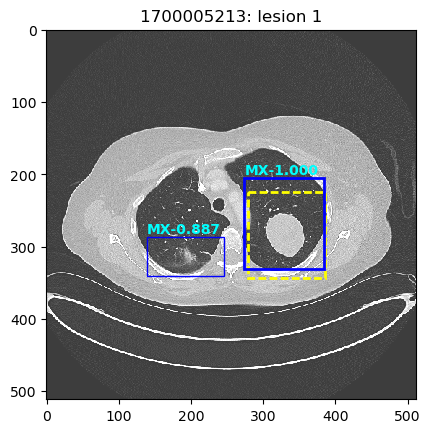}}
	\subfigure[]{\includegraphics[width = 0.300\textwidth, trim=38 30 0 15mm, clip]{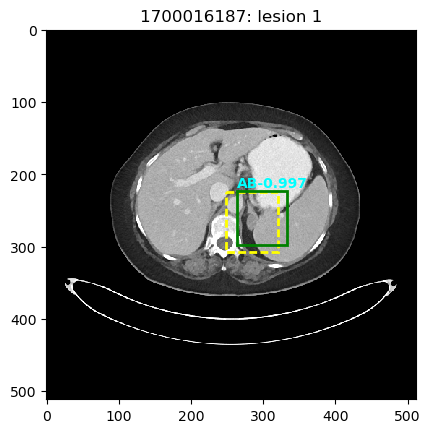}}
	\subfigure[]{\includegraphics[width = 0.300\textwidth, trim=38 30 0 15mm, clip]{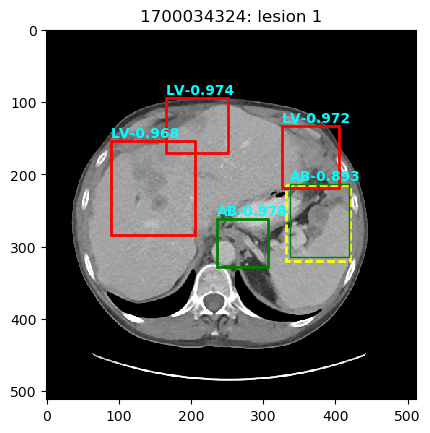}}
	\subfigure[]{\includegraphics[width = 0.300\textwidth, trim=38 30 0 15mm, clip]{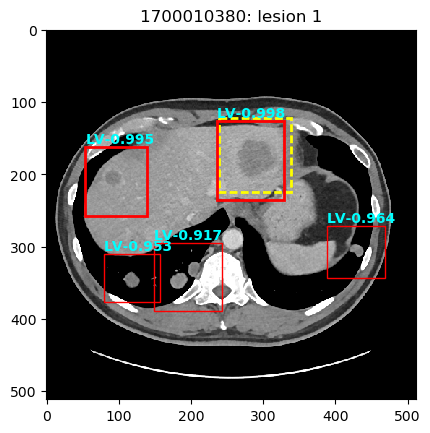}}
	\caption{Six sample detection results are illustrated with the annotation lesion patches as yellow dashed boxes. The outputs from our proposed detection framework are shown in colored boxes with LiVer lesion (LV) in Red, Lung Nodule (LN) in Orange, ABdomen lesion (AB) in Green, Chest Lymph node (CL) in magenta and other MiXed lesions (MX) in blue. (a) Four lung lesions are all correctly detected; (b) Two lymph nodes in mediastinum is presented; (c) A Ground Glass Opacity (GGO) and a mass are detected in the lung; (d) An adrenal nodule; (e) Correct detections on both the small abdomen lymph node nearly aorta but also other metastases in liver and spleen; (f) Two liver metastasis are correctly detected. Three lung metastases are detected but erroneously classified as liver lesions . 
	}
	\label{fig:det_results} \vspace{-4mm}
\end{figure}

Our {\bf lesion categorization} method in Sec. \ref{annot} partitions all lesion patches into five classes $k=5$. After visual inspection supervised by a board-certificated radiologist, four common lesion categories are found, namely lung nodule/lesion, liver lesion, chest lymph nodes and abdominal lesions (mainly kidney lesions and lymph nodes), with high purity scores in each category (0.980 for Lung Nodule, 0.955 for Chest Lymph Node, 0.805 for Liver Lesion and 0.995 for Abdomen Lesion). The per-cluster purity scores are estimated through a visual assessment by an experienced radiologist using a set of 800 randomly selected lesion patches (200 images per-category). The remaining bookmarked annotations are treated as a ``noisy" mixed lesion class. Our optimization framework converges after six iterations, with a high purity score of 92.3\% returned when assessing the statistical similarity or stability of two last iterations. Meanwhile, the top-1 classification accuracy reaches 91.6\%, and later fluctuates by $\pm2\%$ . 

For {\bf lesion detection}, all bookmarked images are divided into training (70\%), validation (15\%), and testing (15\%) sets, by random splitting the dataset at the patient level. Although different lesion types may be present in an image, only one clinical annotation per image is available. We adopt a straightforward evaluation criterion: 1) we take the top one detected lesion candidate box (with the highest detection confidence score) per testing image as the detection result; 2) if the intersection-over-union (IoU) between this predicted box and the ground-truth clinical annotation box is larger than 0.5 (as suggested by the PASCAL criterion \cite{Everingham2015}), the detection is regarded as correct, and vice versa. The lesion category is not considered in this criterion. We denote this evaluation metric as detection accuracy. 
The proposed multi-class lesion detector merely requires 88 ms to process a 512$ \times $512 test bookmarked image on a Titan X Maxwell GPU.

\begin{figure}[t]
	\centering
	\subfigure[]{\includegraphics[width=0.49\linewidth]{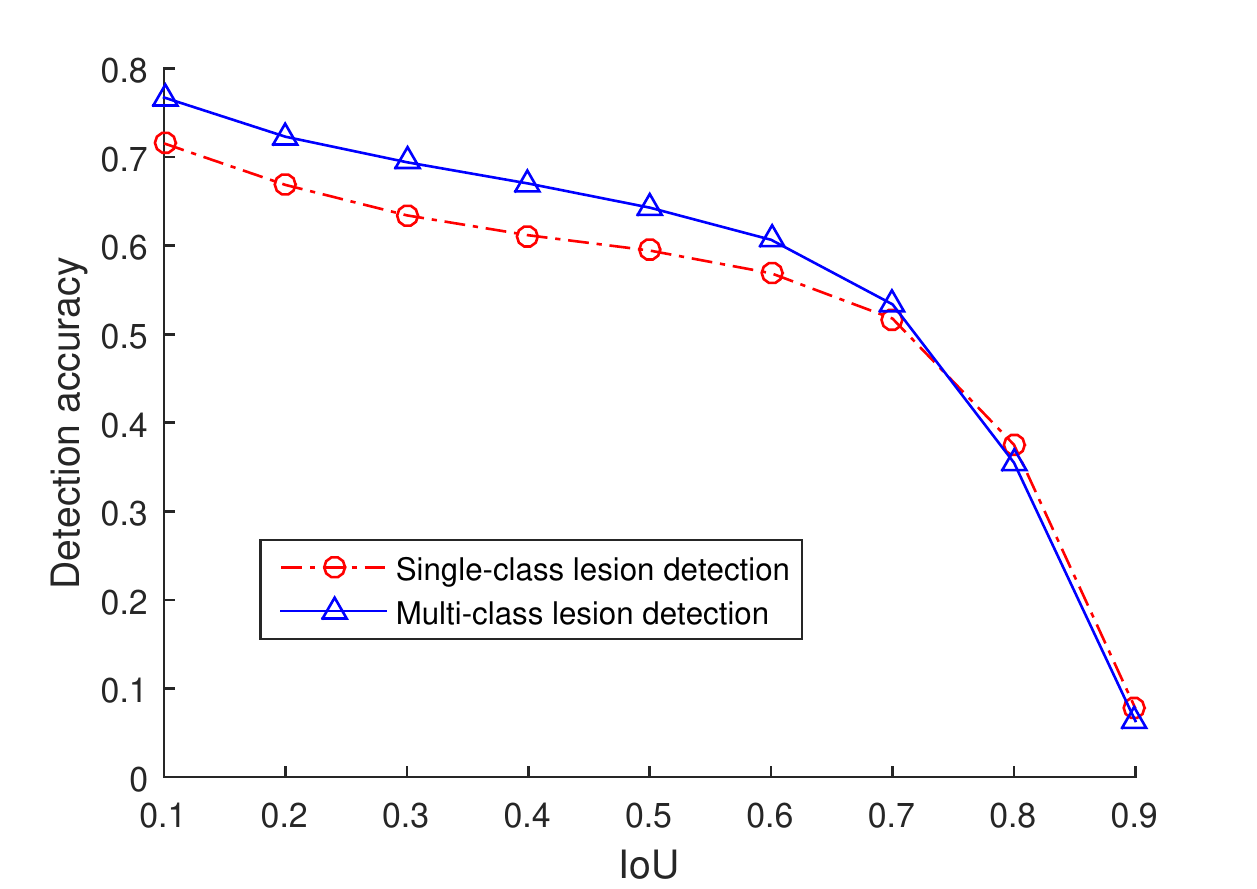}}
	\subfigure[]{\includegraphics[width=0.49\linewidth]{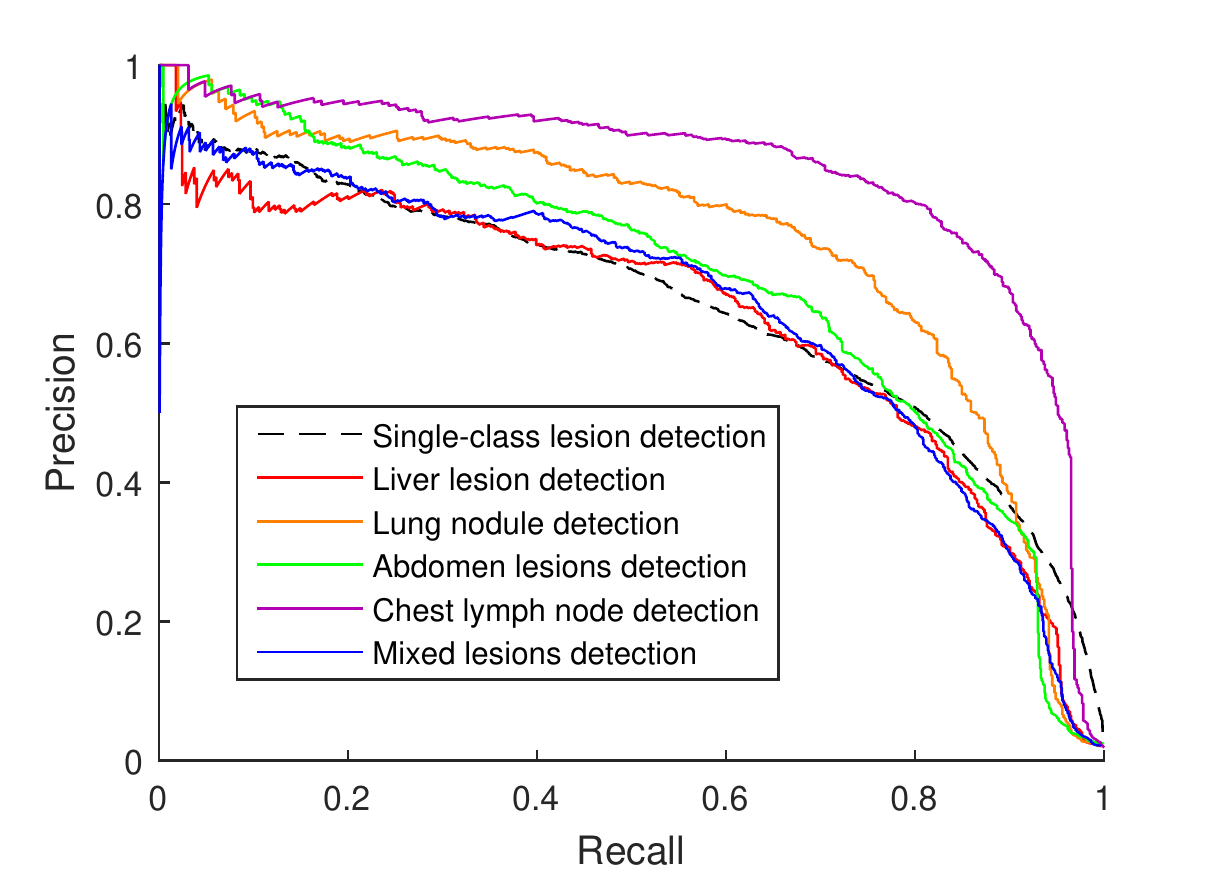}}
	\caption{(a): Detection accuracy curves with different intersection-over-union (IoU) thresholds. (b): Precision-Recall curves of single-class and five category lesion detection when IoU=0.5.}
	\label{fig:iou_curve} \vspace{-3mm}
\end{figure}

\begin{table}[t]
	\centering
	\setlength{\tabcolsep}{5pt}
	\renewcommand{\arraystretch}{1.2}
	\begin{tabular}{p{3.2cm}p{0.9cm}p{1.cm}p{1.2cm}p{1.5cm}p{1.cm}p{1cm}}
		\hline 
		Cluster & 1	& 2	& 3	& 4	& 5	& Overall \\
		\hline
		Cluster category 
		& Liver lesion	& Lung nodule	& abdomen lesions & Chest lymph node & Mixed lesions & \\
		\hline
		Cluster size	& 774	& 837	& 1270	& 860 & 1292	& 5033	\\
		\hline
		Averaged Accuracy: w/o categorization labels (\%)	& 56.04	& 73.30	& 48.79	& 70.77	& 54.85	& 59.45	\\	
		\hline	
		Averaged accuracy: with categorization labels (\%)	& 60.59	& 76.34	& 56.22	& 76.28	& 58.67	& 64.30	\\
		\hline
	\end{tabular}
	\caption{Test detection accuracies of five deep-learned pseudo lesion categories. Note that in DeepLesion dataset, available clinical annotations are only partial (i.e., missing lesion labels). Hence the actual detection accuracies in both configurations are significantly higher than the above reported values since many ``false positive'' detections are later verified to be true lesions.}
	\label{tbl:acc} \vspace{-2mm}
\end{table}

Two lesion detection setups or configurations are examined: single-class (all annotation bounding boxes are considered as one abnormality class), and multi-class detection (with pseudo-category labels). Some illustrative results from the multi-category lesion detector on the testing set are shown in Fig. \ref{fig:det_results}. It can be found that our developed detector is able to detect all five types of lesions and simultaneously provide the corresponding lesion category labels. Furthermore, some detection boxes currently considered as false alarms actually belong to true lesions because the lesions bounding boxes are only partially labeled by clinical bookmark annotations. 
Detailed statistics of the five deeply discovered lesion clusters in the test set are provided in Table \ref{tbl:acc}. This outlines the types of lesions in the clusters that have been verified by radiologists. The averaged accuracy of the single-class detection is 59.45\% (testing) and this score becomes 64.3\% for multi-class detection (testing). From Table \ref{tbl:acc}, the multi-category detector also demonstrates accuracy improvements of 3$\sim$8\% per lesion cluster or category compared against the one-class abnormality detector. Single-class abnormality detection appears to be a more challenging task since it tackles detecting various types of lesions at once. This validates that better lesion detection models can be trained if we can perform unsupervised lesion categorization from a large collection of retrospective clinical data. 

The default IoU threshold is set as 0.5. Fig. \ref{fig:iou_curve} (a) illustrates the detection accuracy curves of both detection models under different IoU thresholds. The multi-category lesion detection achieves the better overall accuracy while being able to assign the lesion labels at the same time. Fig. \ref{fig:iou_curve} (b) shows the corresponding detection precision-recall curves. The performances of lung lesion detection and chest lymph node detection significantly outperform the one-class abnormality detection.

\section{Conclusion}
\label{conclusion}
In this paper, we mine, categorize and detect one type of clinical annotations stored
in the hospital PACS system as a rich retrospective data source, to build a large-scale
Radiology lesion image database. We demonstrate the strong feasibility of employing
a new multi-category lesion detection paradigm via unified deep categorization and detection.
Highly promising lesion categorization and detection performances, based on
the proposed dataset, are reported. To the best of our knowledge, this work is the first
attempt of building a scalable, multi-purpose CAD system using abundant retrospective
medical data. This is done almost effortlessly since no new arduous image annotation
workload is necessary. Our future work include extending bookmarked images to incorporate
their successive slices for the scalable and precise lesion volume measurement;
extracting and integrating the lesion diagnosis prior from radiology text reports, and improved multi-category detection methods.

\bibliographystyle{abbrv}
\bibliography{miccai2017}

\begin{thebibliography}{10}

\bibitem{Green16Liv}
A.~Ben-Cohen, I.~Diamant, E.~Klang, M.~Amitai, and H.~Greenspan.
\newblock Fully convolutional network for liver segmentation and lesions
  detection.
\newblock In {\em MICCAI LABELS-DLMIA}, 2016.

\bibitem{deng2009imagenet}
J.~Deng, W.~Dong, R.~Socher, L.-J. Li, K.~Li, and L.~Fei-Fei.
\newblock Imagenet: A large-scale hierarchical image database.
\newblock In {\em IEEE CVPR}, pages 248--255, 2009.

\bibitem{esteva2017dermatologist}
A.~Esteva, B.~Kuprel, R.~A. Novoa, J.~Ko, S.~M. Swetter, H.~M. Blau, and
  S.~Thrun.
\newblock Dermatologist-level classification of skin cancer with deep neural
  networks.
\newblock {\em Nature}, 542(7639):115--118, 2017.

\bibitem{Everingham2015}
M.~Everingham, A.~Eslami, L.~{Van Gool}, C.~Williams, J.~Winn, and
  A.~Zisserman.
\newblock The pascal visual object classes challenge: A retrospective.
\newblock {\em Int. J. Comp. Vis.}, 111(1):98--136, 2015.

\bibitem{Gomes2010}
R.~Gomes, A.~Krause, and P.~Perona.
\newblock Discriminative clustering by regularized information maximization.
\newblock In {\em NIPS}, pages 775--783, 2010.

\bibitem{lin2014microsoft}
T.-Y. Lin, M.~Maire, S.~Belongie, J.~Hays, P.~Perona, D.~Ramanan,
  P.~Doll{\'a}r, and C.~L. Zitnick.
\newblock Microsoft coco: Common objects in context.
\newblock In {\em ECCV}, pages 740--755, 2014.

\bibitem{Liu16Rcnn}
J.~Liu, D.~Wang, Z.~Wei, L.~Lu, L.~Kim, E.~Turkbey, and R.~M. Summers.
\newblock Colitis detection on computed tomography using regional convolutional
  neural networks.
\newblock In {\em IEEE ISBI}, 2016.

\bibitem{Ren15Faster}
S.~Ren, K.~He, R.~Girshick, and J.~Sun.
\newblock Faster r-cnn: Towards real-time object detection with region proposal
  networks.
\newblock In {\em NIPS}, pages 91--99, 2015.

\bibitem{Sim15Vgg}
K.~Simonyan and A.~Zisserman.
\newblock Very deep convolutional networks for large-scale image recognition.
\newblock In {\em ICLR}. arxiv.org/abs/1409.1556, 2015.

\bibitem{Taj15Pulmonary}
N.~Tajbakhsh, M.~B. Gotway, and J.~Liang.
\newblock Computer-aided pulmonary embolism detection using a novel
  vessel-aligned multi-planar image representation and convolutional neural
  networks.
\newblock In {\em MICCAI}, pages 62--69. Springer, 2015.

\end{thebibliography}
\end{document}